\documentclass{article}

     \PassOptionsToPackage{numbers}{natbib}

\usepackage[]{arxiv}
\usepackage{environ}
\newcommand{\acksection}{\section*{Acknowledgments and Disclosure of Funding}}
\NewEnviron{ack}{%
  \acksection
  \BODY
}

\usepackage[utf8]{inputenc} 
\usepackage[T1]{fontenc}    
\usepackage{hyperref}       
\usepackage{url}            
\usepackage{booktabs}       
\usepackage{amsfonts}       
\usepackage{nicefrac}       
\usepackage{microtype}      
\usepackage{xcolor}         

\usepackage{gensymb}
\usepackage{amsmath}
\usepackage{graphicx}
\usepackage{textcomp}
\usepackage{amsthm}
\newtheorem{theorem}{Theorem}

\newtheorem{corollary}{Corollary}

\usepackage{subcaption}     
\usepackage{verbatim}       
\usepackage{adjustbox}      
\usepackage{multirow, tabularx, array}        

\title{Do highly over-parameterized neural networks generalize since bad solutions are rare?}

\author{%
  Julius Martinetz\\
  Machine Learning Group\\
  Technical University Berlin\\
  Berlin, Germany \\
  \texttt{j.martinetz@tu-berlin.de} \\
  \And
  Thomas Martinetz\\
  Institute for Neuro- and Biocomputing\\
  University of L\"ubeck\\
  L\"ubeck, Germany \\
  \texttt{martinetz@inb.uni-luebeck.de} \\
}

\begin{document}

\maketitle

\begin{abstract}
 We study over-parameterized classifiers where Empirical Risk Minimization (ERM) for learning leads to zero training error. In these over-parameterized settings there are many global minima with zero training error, some of which generalize better than others. We show that under certain conditions the fraction of "bad" global minima with a true error larger than $\varepsilon$ decays to zero exponentially fast with the number of training data $n$. The bound depends on the distribution of the true error over the set of classifier functions used for the given classification problem, and does not necessarily depend on the size or complexity (e.g.\ the number of parameters) of the classifier function set. This insight may provide a novel perspective on the unexpectedly good generalization even of highly over-parameterized neural networks.
 We substantiate our theoretical findings through experiments on synthetic data and a subset of MNIST. Additionally, we assess our hypothesis using VGG19 and ResNet18 on a subset of Caltech101.
\end{abstract}

\section{Introduction}
Extreme upscaling and deepening of neural networks has led to a quantum leap in many real-world object recognition tasks \cite{KrSuHi2012}. The same is true for neural networks in Natural Language Processing \cite{Bert2018} and reinforcement learning \cite{DeepMindAtari2015}. By enlarging the networks appropriately, the level of performance increases, and even entering regimes with many more network parameters than training samples does not harm generalization. This is usually attributed to an appropriate regularization. 
However, generalization is also observed in highly over-parameterized regimes and without explicit regularization, where even random label assignments or even random image data can be memorized~\cite{ZhangBengio2017}. Belkin called this the "modern" interpolation regime' \cite{Belkin2019}, although this effect has been studied for some time \cite{LoogTax2020}.

The generalization capabilities of large neural networks do not seem to be harmed by the millions of parameters in today's popular architectures. Due to their large, inherent capacity the training error is zero after training by Empirical Risk Minimization (ERM), i.e.\ the training data is memorized. Nevertheless, the test error can be unexpectedly low \cite{ZhangBengio2017}, even after only a few handful of training samples \cite{linse_large_2023}. This is in contrast to traditional machine learning wisdom, where one would not expect any generalization in these highly over-parameterized regimes, where the data is highly over-fitted. Statistical learning theory based on uniform generalization bounds requires a high probability that there is no solution with a large deviation between empirical and expected risk, i.e. the training and the true error. However, this is not the case in over-parameterized regimes. There is no uniform generalization bound. In these highly over-parameterized regimes always also "bad" solutions with zero training but large true error do exist. Interestingly, they hardly seem to occur in practice.

In this paper, we investigate over-parameterized classifiers with training error zero solutions. A classifier receives inputs $x$ from an input distribution $P(x)$ and assigns a class label using a classifier function $h(x)$. The input $x$ has a true label $y$ with probability $P(y\vert x)$. For each input $x$, the classifier produces a loss $L(y, h(x))\in \{0,1\}$, $0$ if the classification is correct and $1$ if the classification is incorrect. The error of the classifier $h$ on the whole data distribution $P(x,y)=P(y\vert x)P(x)$ is given by the expected risk
\begin{equation*}
    E(h)=\int  L(y,h(x)) P(x,y) \,dx\, dy. 
\end{equation*}
$E(h)$ is also called true error and always lies between $0$ and $1$. 
The classifier or predictor $h$ is usually chosen from a function set $\cal H$. For example, $\cal H$ is determined by the architecture of a neural network, and the allocated $h$ is determined by the network parameters. Learning means selecting an $h$ from the set $\cal H$ so as to minimize $E(h)$. This is often done by Empirical Risk Minimization (ERM). Let ${\cal S}=\{(x_1,y_1),... ,(x_n,y_n)\}$ be a so-called training set of $n$ input samples $x$ together with their labels $y$ which are drawn i.i.d.\ from $P(x,y)$. The empirical risk is defined as the average loss on this training set  
\begin{equation*}
    E_{{\cal S}}(h)=\frac{1}{n}\sum_{i=1}^n L(y_i,h(x_i)) 
\end{equation*}
and is also called training error. Learning via ERM chooses an $h\in \cal H$ that minimizes $E_{{\cal S}}(h)$. However, $E(h)$ and $E_{{\cal S}}(h)$ may deviate. This deviation is called the generalization gap. 

If the generalization gap is small, a small training error provides a good solution with a small true error. Statistical learning theory based on VC-dimension \cite{Vapnik1998} or Rademacher complexity \cite{BartlettM2002} provides uniform bounds for the generalization gap. The generalization gap becomes small with increasing $n$ for each $h\in \cal H$. The objective is to completely rule out "bad" solutions with large generalization gaps, at least with high probability. However, these bounds do not exist in highly over-parameterized regimes. There are always $h\in \cal H$ left with large generalization gaps. Nevertheless, in practice learning takes place and ERM provides good solutions that generalize well also in these over-parameterized regimes. There have been theoretical and empirical attempts to understand this "mystery". It has been argued, that this might be due to implicit regularization of (stochastic) gradient descent \cite{Neyshabur2017}, \cite{brutzkus2018}, \cite{Soudry2018}, \cite{Lyu2020}, \cite{advani2020}, \cite{Vardi2023}. Further, a number of novel algorithmic-dependent uniform generalization bounds based on norm or compression measures have been supposed for explanation \cite{NeyshaburColt2015}, \cite{Bartlett2017}, \cite{Golowich2018}, \cite{sanjeev2018}, \cite{LiLiang2018}, \cite{Kolter2019}, \cite{neyshabur2019}, \cite{Wei2019}, \cite{LiangPoggio2019}. The Neural Tangent Kernel (NTK) with its linearization at initialization gives further insights for very wide neural networks \cite{Jacot2018}, \cite{Chizat2018O}, \cite{AllenZhu2018LearningAG}, \cite{Arora2019OnEC}, \cite{Arora2019FineGrainedAO}, \cite{Sohl-Dickstein2020}, \cite{Min2021}. Another approach applies the concepts of algorithmic stability \cite{hardt16}, \cite{mou18a}, \cite{lei2022}, \cite{Oneto2023}.
But still there is empirically based scepticism that these bounds already help \cite{Jiang2020fantastic} or that uniform convergence bounds are fundamentally the right approach \cite{Nagarajan2019UniformCM}.  

As "bad" solutions cannot be ruled out in over-parameterized regimes, we conjecture that within the set of ERM solutions, the proportion of "bad" classifiers nevertheless remains small. This hypotheses offers a potential alternative explanation for the enigma of good generalization in over-parameterized regimes and corresponds to and supports the empirical studies in \cite{Goldstein2023loss}. If the fraction of "bad" classifiers is small among the global minima with zero training error, the ERM algorithm is more likely to converge to a favorable solution rather than an unfavorable one. This perspective introduces a novel angle, challenging the conventional wisdom in machine learning that typically assumes the prevalence of "bad" solutions in highly over-parameterized regimes, and that it needs appropriate regularization to converge to good solutions. 

\section{Preliminaries, Notations and Definitions}

We assume ${\cal H}={\cal H}_{\cal W}$ to be parameterized on a compact set ${\cal W}\subseteq \mathbb{R}^N$ with non-zero Lebesgue measure, and each $w\in {\cal W}$ determines a classifier function $h_w\in {\cal H}_{\cal W}$. If a neural network is used as classifier, then $w$ are the weights of the neural network. For each $w$ we obtain a true error $E(h_w)=E(w)$. We assume $E(w)$ to be integrable on ${\cal W}$ and ${\cal H}_{\cal W}$ to have a large but finite VC-dimension. $E_{min}=\min_{w\in{\cal W}} E(w)$ denotes the minimum true error of the classifier function set ${\cal H}_{\cal W}$ on the given classification problem. We work with the following subsets of ${\cal W}$:
\begin{eqnarray*}
    {\cal W}_\varepsilon&:& \hbox{set of all $w\in{\cal W}$ with $E(w)\geq E_{min}+\varepsilon$}\quad (\varepsilon\geq 0)\\
    {\cal W}({\cal S})&:& \hbox{set of all $w\in{\cal W}$ with $E_{{\cal S}}(w)=0$ }\\
    {\cal W}_\varepsilon({\cal S})&:& \hbox{set of all  $w\in{\cal W}_\varepsilon$ with $E_{{\cal S}}(w)=0$ }
\end{eqnarray*}
${\cal W}({\cal S})$ is the set of all global minima on the training error $E_{\cal S}(w)$ for a given training set ${\cal S}$. Since we are interested in the over-parameterized regime, i.e.\ the training set size $n=\vert {\cal S}\vert$ is small compared to the complexity of the classifier, 
we assume ${\cal W}({\cal S})$ to be of non-zero measure for almost all ${\cal S}$ with $\vert {\cal S}\vert=n$. We can take it as a definition for over-parameterization. Hence, the ERM algorithm ends up in one of these global minima\footnote{Convergence to global minima of ERM algorithms such as gradient descent or its variants is a topic on its own rights. However, in highly over-parameterized settings convergence to zero training error is usually not a problem.}. ${\cal W}_\varepsilon({\cal S})$ is what we call the set of "bad" global minima with true errors larger than $E_{min}+\varepsilon$. We have ${\cal W}_\varepsilon({\cal S})\subseteq{\cal W}({\cal S})\subseteq{\cal W}$.
Further, we use the notations $\Omega=\vert {\cal W}\vert$, $\Omega_\varepsilon=\vert {\cal W}_\varepsilon\vert$, $\omega({\cal S})=\vert {\cal W}({\cal S})\vert$ and $\omega_\varepsilon({\cal S})=\vert {\cal W}_\varepsilon({\cal S})\vert$ for the size (volume) of these parameter sets. In the following the fractions 
\begin{eqnarray*}
    g_\varepsilon=\frac{\Omega-\Omega_\varepsilon}{\Omega}&:& \hbox{fraction of "good" classifiers within ${\cal W}$ }\\
    \phi_\varepsilon({\cal S})=\frac{\omega_\varepsilon({\cal S})}{\omega({\cal S})}&:& \hbox{fraction of "bad" classifiers within the set of global minima ${\cal W}({\cal S})$}
\end{eqnarray*}
are important. $g_\varepsilon$ as the fraction of "good" classifiers within ${\cal W}$ is a measure for the "appropriateness" of ${\cal H}$ for the given classification problem. It is a measure for the bias of the classifier function set towards the given classification problem. $\phi_\varepsilon({\cal S})$ is the main property we are looking at. If $\phi_\varepsilon({\cal S})$ is small, the ERM algorithm should more likely end up in a good solution. We show that under certain conditions $\phi_\varepsilon({\cal S})$ is very likely to become small with increasing $n$ and relate it to $g_\varepsilon$. 

\subsection{Density of Classifiers (DOC)}

For quantifying the parameter volume of good and bad classifiers we 
introduce the "density of classifiers" $D(E)$ at true error $E$ defined such that
\begin{equation*}
    \Omega_\varepsilon=\int_{E_{min}+\varepsilon}^1  D(E)\, dE \qquad \hbox{for each} \qquad 0\leq \varepsilon\leq 1-E_{min}.
\end{equation*}
$D(E)dE$ is the volume of parameters $w\in {\cal W}$ with $E\leq E(w) \leq E+dE$. For $0\leq E < E_{min}$ we set $D(E)=0$. 

\begin{figure}[t]
	\centering
	\begin{subfigure}[t]{0.48\textwidth}
		\includegraphics[width=0.99\textwidth]{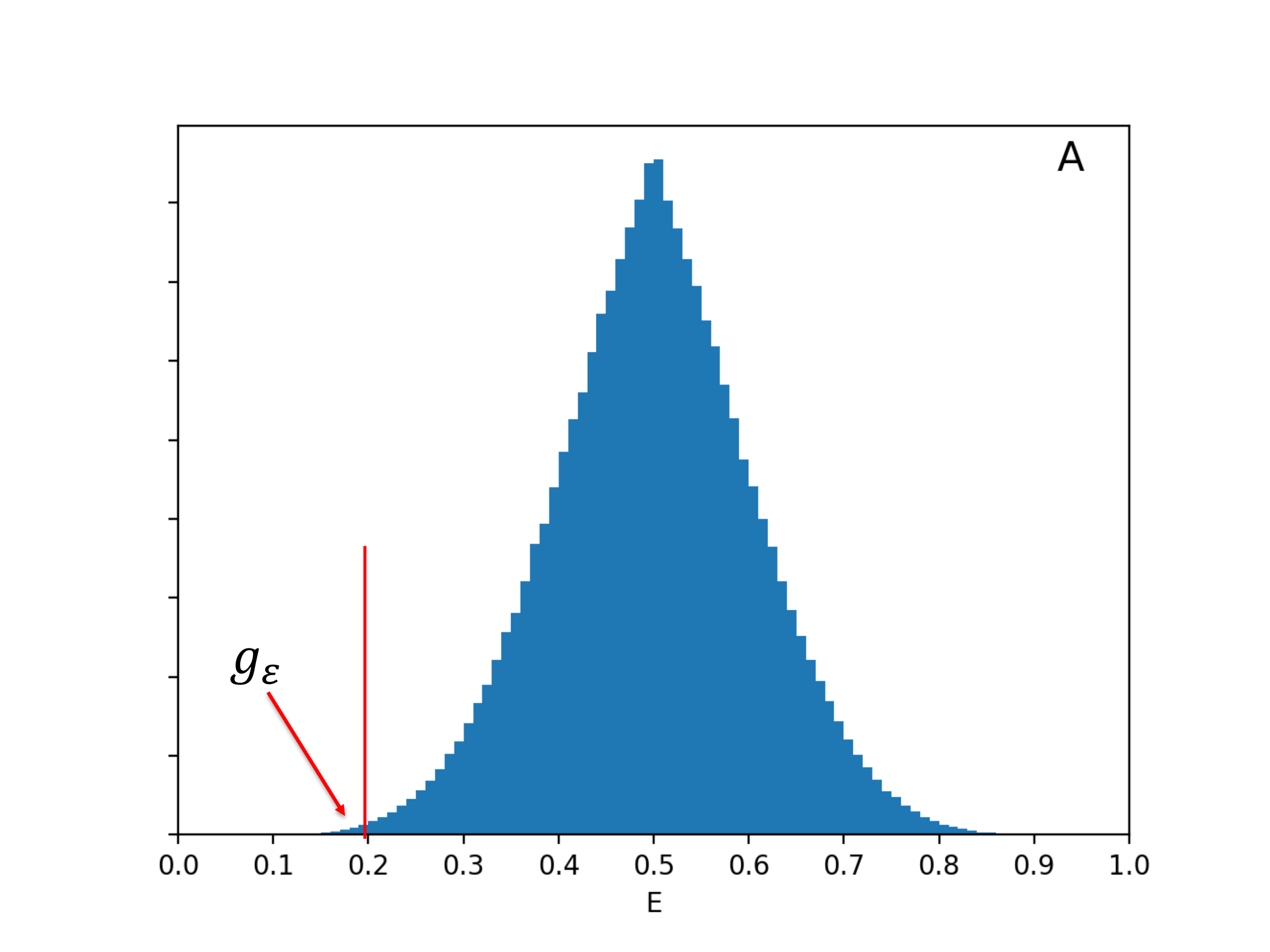}
	\end{subfigure}
    ~
	\begin{subfigure}[t]{0.48\textwidth}
		\includegraphics[width=0.99\textwidth]{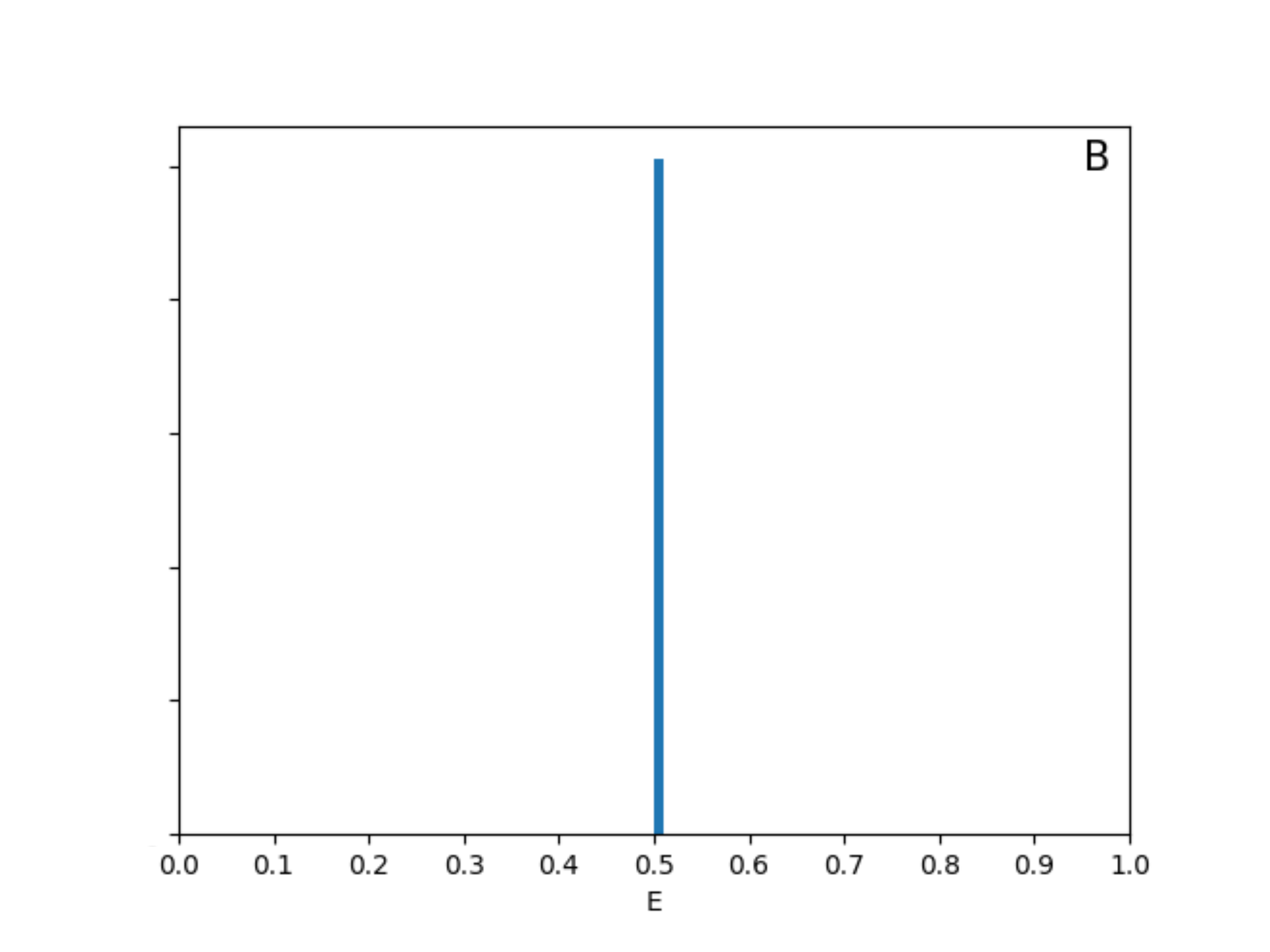}
	\end{subfigure}

	\caption{The density of dlassifiers (DOC) for a neural network from our experiments (A) and a binary classification problem with random labels (B). In (A) the DOC is shown as a histogram of the true error for one million randomly chosen network weights. $g_\varepsilon$ gives the fraction of weights with true errors lower than $0.2$. In a binary classification problem with random labels (B), for each classifier $w$ the true error $E(w)$ is always $0.5$ and, hence, the DOC is a peak at $E=0.5$.}
\label{fig:DOC}
\end{figure}

In Fig.~\ref{fig:DOC} we illustrate the density of classifiers (DOC) for two scenarios. On the left (A) we see the DOC of a neural network from our experiment in Section~4.1. For one million network weights $w\in{\cal W}$ uniformly chosen from ${\cal W}$ their corresponding true error $E(w)$ is determined (details in Section~4.1). The DOC in A shows the distribution of the true errors. The part left of the red line illustrates $g_\varepsilon=\int_0^{E_{min}+\varepsilon} D(E)\, dE/\Omega$, the fraction of good classifiers with a true error smaller than $E_{min}+\varepsilon$.
On the right (B) we show the DOC in case of a binary classification problem with random class labels (each class with equal probability). For each $w$, the true error is $E(w)=0.5$ and, hence, the DOC is a peak at $E=0.5$. 

\section{Results}

For ${\cal S}$ randomly drawn from $P(x,y)^n$, the parameter set size $\omega_\varepsilon({\cal S})$ of "bad" global minima is a random variable with $0\leq \omega_\varepsilon({\cal S}) \leq \Omega$. The first and main Theorem, for which the proof is given in the Appendix, looks at its mean value: 
\begin{theorem}
    For each $0\leq \varepsilon\leq 1-E_{min}$ 
    \begin{eqnarray}
        \langle \omega_\varepsilon({\cal S})\rangle_{\cal S} &=&\int_{{\cal W_\varepsilon}} (1-E(w))^n \, dw\label{AvSetSize}\\
        &=& \int_{E_{min}+\varepsilon}^1  (1-E)^n D(E)\, dE\nonumber\\
        &\leq& \Omega_\varepsilon (1-(E_{min}+\varepsilon))^n\nonumber.
    \end{eqnarray}
\end{theorem}
Theorem 1 tells us, that the mean parameter set size (parameter volume) of "bad" global minima decreases to zero exponentially fast with increasing training set size $n$. However, this might not suffice for $\phi_\varepsilon({\cal S})$ to become small, since the mean parameter set size (volume) of ALL global minima, $\langle \omega({\cal S})\rangle_{\cal S}$, decreases exponentially fast as well. $\langle \omega({\cal S})\rangle_{\cal S}$ is given by \eqref{AvSetSize} for $\varepsilon=0$. Either it converges to zero, e.g.\ if $E_{min}>0$, or to the mean set size of solutions with true error zero in case $E_{min}=0$. If the classification problem is separable, this set size can be of non-zero measure. 

However, the mean parameter set size of ALL global minima converges much slower than the mean set size of "bad" global minima and, hence, the mean set size of "bad" global minima will become very small relative to the mean set size of all global minima. The following corollary quantifies it:
\begin{corollary}
    For each $0\leq\varepsilon< 1-E_{min}$
    \begin{eqnarray}
        \frac{\langle \omega_\varepsilon({\cal S})\rangle_{\cal S} }{\langle \omega({\cal S})\rangle_{\cal S}}&=&\frac{\int_{E_{min}+\varepsilon}^1(1-E)^n D(E)\, dE}{\int_0^1(1-E)^n D(E)\, dE}\label{RelFracDOC}\\
            &\leq& \frac{\Omega_\varepsilon}{\Omega}\frac{1}{(1-g_{\varepsilon/2})+g_{\varepsilon/2}\;e^{\frac{\varepsilon}{2} n}}\label{eq:leq_one}\\
            &\leq&\frac{1}{g_{\varepsilon/2}}\;e^{-\frac{\varepsilon}{2} n} \qquad \hbox{if} \quad g_{\varepsilon/2}>0.\label{ExpDecay}
\end{eqnarray}
\end{corollary}
The mean set size of "bad" global minima relative to mean set size of all global minima diminishes at least exponentially fast with $n$. Note, that the prefactor in \eqref{ExpDecay} is determined by the inverse of the fraction of "good" classifiers within ${\cal W}$ (better than $\varepsilon/2$). The larger the fraction of good classifiers to begin with, i.e., the better the bias of the classifier set, the smaller the prefactor. It does not depend on the size of ${\cal W}$ and does not increase necessarily with $\Omega$, e.g., with the number of parameters. We will demonstrate this later in our experiments. The exponent is linear in $\varepsilon$. 

Corollary 1 tells us, that the mean set size of "bad" global minima is small relative to the mean set size of all global minima. In a concrete setting with a given training set ${\cal S}$, $\phi_\varepsilon({\cal S})=\omega_\varepsilon({\cal S})/\omega({\cal S})$, i.e., the given fraction of "bad" global minima should be small relative to the given fraction of all global minima. In general    
    \begin{equation*}
        \langle \phi_\varepsilon({\cal S})\rangle_{\cal S} =\left\langle\frac{\omega_\varepsilon({\cal S})}{\omega({\cal S})}\right\rangle_{\cal S} 
        \neq \frac{\langle \omega_\varepsilon({\cal S})\rangle_{\cal S} }{\langle \omega({\cal S})\rangle_{\cal S} },
    \end{equation*}
but with the following corollary we get a condition for the l.h.s.\ to be smaller than the r.h.s., which would be sufficient:
\begin{corollary}
If and only if $\phi_\varepsilon({\cal S})$ and $\omega({\cal S})$ do not correlate negatively, then
    \begin{eqnarray}
    \langle \phi_\varepsilon({\cal S})\rangle_{\cal S}
        &\leq& \frac{\langle \omega_\varepsilon({\cal S})\rangle_{\cal S}}{\langle \omega({\cal S})\rangle_{\cal S} }.
         \label{eq:theorem2}\nonumber\\
    &\leq&\frac{\int_{E_{min}+\varepsilon}^1(1-E)^n D(E)\, dE}{\int_0^1(1-E)^n D(E)\, dE}.
    \label{eq:frac_ineq}
    \end{eqnarray}
    Equality is given if and only if $\phi_\varepsilon({\cal S})$ and $\omega({\cal S})$ do not correlate.
\end{corollary}
Corollary 2 directly follows from its condition, since then $\left\langle \phi_\varepsilon({\cal S})\omega({\cal S})\right\rangle_{\cal S}-\left\langle \phi_\varepsilon({\cal S})\right\rangle_{\cal S}\left\langle \omega({\cal S})\right\rangle_{\cal S}\geq 0 $, and from the definition of $\phi_\varepsilon({\cal S})$ we have $\phi_\varepsilon({\cal S})\omega({\cal S})= \omega_\varepsilon({\cal S})$.

With the Markov-inequality the probability, that $\phi_\varepsilon({\cal S})$ is larger than a $\gamma>0$, is given by
\begin{equation*}
    Prob\left\{\phi_\varepsilon({\cal S}) \geq \gamma \right\} \leq \frac{1}{\gamma}\frac{\int_{E_{min}+\varepsilon}^1(1-E)^n D(E)\, dE}{\int_0^1(1-E)^n D(E)\, dE}\leq \frac{1}{\gamma g_{\varepsilon/2}}\;e^{-\frac{\varepsilon}{2} n}.\label{ProbExpDecay}
\end{equation*}
The probability, that the fraction $\phi_\varepsilon({\cal S})$ of "bad" ERM solutions within the set of all ERM solutions is not small (smaller than $\gamma$) for a given ${\cal S}$, decreases exponentially fast to zero with the number of training data $n$. "Bad" solutions become rare with high probability. It is not the absolute number (density) of classifiers, which determines the r.h.s, but only the shape of $D(E)$, i.e., the normalized $D(E)$. The shape of $D(E)$ does not necessarily depend on the number of parameters of our classifier functions. 

\subsection{Mean true error}

In the following we derive an expression for the mean true error $E_n$ over all global minima ${\cal W}_{\cal S}$ and all ${\cal S}$. For a given $\cal S$ we define $Q_{\cal S}(E)$ such that
\begin{equation}
    \phi_\varepsilon({\cal S})=\frac{\omega_\varepsilon({\cal S})}{\omega({\cal S})}=\int_{E_{min}+\varepsilon}^1  Q_{\cal S}(E)\, dE \qquad \hbox{for each} \qquad 0\leq \varepsilon\leq 1-E_{min}.\label{def:Q_S}
\end{equation}
$Q_{\cal S}(E)$ is the normalized density of global minima at true error $E$ for the given $\cal S$. One will rather end up with a classifier at true errors where the density $Q_{\cal S}(E)$ is large. The mean true error $E_{\cal S}$ over all global minima of a given $\cal S$ is  
\begin{equation*}
E_{\cal S}=\int_{0}^1  E\, Q_{\cal S}(E)\, dE.
\end{equation*}
With $Q_n(E)=\langle Q_{\cal S}(E)\rangle_{\cal S}$ we obtain from \eqref{def:Q_S}
\begin{equation}
    \langle \phi_\varepsilon({\cal S})\rangle_{\cal S}=\int_{E_{min}+\varepsilon}^1  Q_n(E)\, dE \qquad \hbox{for each} \qquad 0\leq \varepsilon\leq 1-E_{min},
    \label{def:Q_n}
\end{equation}
and the mean true error $E_n$ over all global minima ${\cal W}_{\cal S}$ and all ${\cal S}$ is accordingly given by 
\begin{equation}
    E_n=\int_{0}^1  E\, Q_n(E)\, dE.
    \label{def:mean_E}
\end{equation}
If (and only if) Corollary 2 with equality in \eqref{eq:frac_ineq} is valid for each $0\leq \varepsilon\leq 1-E_{min}$, then comparing \eqref{eq:frac_ineq} and \eqref{def:Q_n} gives
\begin{equation*}
    Q_n(E) =\frac{(1-E)^n D(E)\,dE}{\int_{0}^1 (1-E)^n D(E)\,dE },
    \label{eq:Q_n}
\end{equation*}
and with \eqref{def:mean_E} we obtain for the mean true error
\begin{equation}
    E_n=\int_{0}^1\frac{E(1-E)^n D(E)}{\int_{0}^1 (1-E)^n D(E)\,dE}\,dE.
    \label{eq:mean_E}
\end{equation}
In this case the mean true error over all ERM solutions is exactly determined by the density of classifiers $D(E)$. Again, it is not the absolute number (density) of classifiers, which determines $E_n$, but only the shape of $D(E)$.

In case that equality in \eqref{eq:frac_ineq} is not valid, we have to extend Corollary 2. For each $\varepsilon, \varepsilon'$ with $0\leq\varepsilon<\varepsilon'\leq 1-E_{min}$ we denote by ${\cal W}_{\varepsilon<\varepsilon'}({\cal S})$ the set of all $w\in{\cal W}({\cal S})$ with $E_{min}+\varepsilon\leq E(w)\leq E_{min}+\varepsilon'$ and with $\omega_{\varepsilon<\varepsilon'}({\cal S})$ its size. Since $\omega_{\varepsilon<\varepsilon'}({\cal S})=\omega_{\varepsilon}({\cal S})-\omega_{\varepsilon'}({\cal S})$ and with \eqref{RelFracDOC} we obtain
\begin{equation*}
       \frac{\langle \omega_{\varepsilon<\varepsilon'}({\cal S})\rangle_{\cal S} }{\langle \omega({\cal S})\rangle_{\cal S}}=\frac{\int_{E_{min}+\varepsilon}^{E_{min}+\varepsilon'}(1-E)^n D(E)\, dE}{\int_0^1(1-E)^n D(E)\, dE}.
\end{equation*}
With $\phi_{\varepsilon<\varepsilon'}({\cal S})=\omega_{\varepsilon<\varepsilon'}({\cal S})/\omega({\cal S})$ and analog to Corollary 2, if and only if $\phi_{\varepsilon<\varepsilon'}({\cal S})$ and $\omega({\cal S})$ do not correlate negatively for each $0\leq\varepsilon<\varepsilon'\leq 1-E_{min}$, then
\begin{equation*}
       \langle \phi_{\varepsilon<\varepsilon'}({\cal S})\rangle_{\cal S}
        \leq\frac{\int_{E_{min}+\varepsilon}^{E_{min}+\varepsilon'}(1-E)^n D(E)\, dE}{\int_0^1(1-E)^n D(E)\, dE}\qquad \hbox{for each} \qquad 0\leq\varepsilon<\varepsilon'\leq 1-E_{min}.
\end{equation*}
Since $\phi_{\varepsilon<\varepsilon'}({\cal S})=\phi_{\varepsilon}({\cal S})-\phi_{\varepsilon'}({\cal S})$ and with \eqref{def:Q_n}
\begin{equation*}
    \langle \phi_{\varepsilon<\varepsilon'}({\cal S})\rangle_{\cal S}=\int_{E_{min}+\varepsilon}^{E_{min}+\varepsilon'}  Q_n(E)\, dE \qquad \hbox{for each} \qquad 0\leq\varepsilon<\varepsilon'\leq 1-E_{min}.
\end{equation*}
But then
\begin{equation*}
    Q_n(E) \leq\frac{(1-E)^n D(E)\,dE}{\int_{0}^1 (1-E)^n D(E)\,dE }
    \label{ineq:Q_n}
\end{equation*}
and with \eqref{def:mean_E}
\begin{equation}
    E_n\leq\int_{0}^1\frac{E(1-E)^n D(E)}{\int_{0}^1 (1-E)^n D(E)\,dE}\,dE.
    \label{ineq:mean_E}
\end{equation}
Interestingly, in our experiments in the next section $E_n$ is not only bounded by \eqref{ineq:mean_E} but in most cases exactly determined by \eqref{eq:mean_E}.

\subsection{Discussion of results}

\begin{figure}[t]
	\centering
		\includegraphics[width=0.8\textwidth]{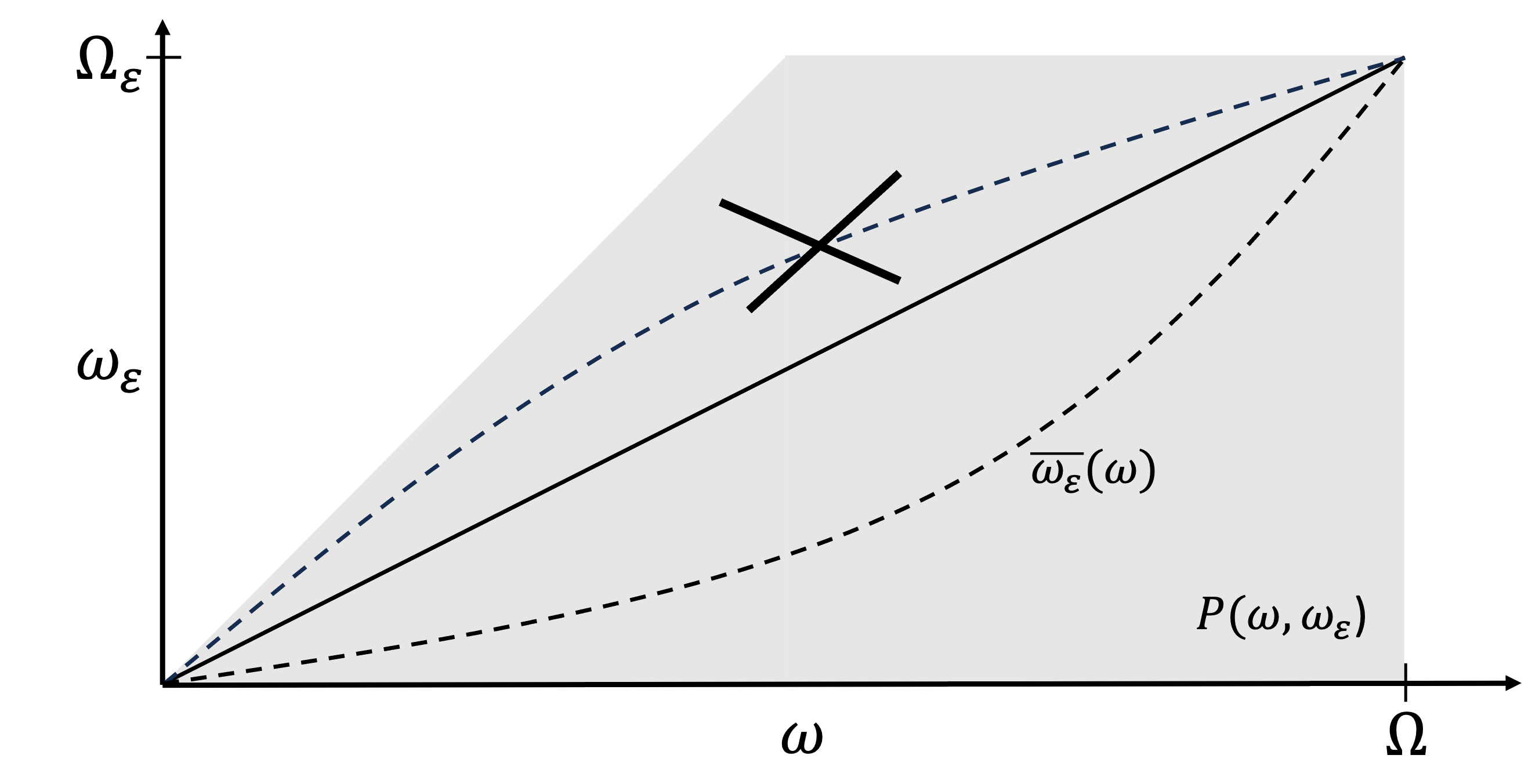}

	\caption{Illustration of the structure of the distribution $P(\omega,\omega_\varepsilon)$ of the two random variables $\omega$ and $\omega_\varepsilon$. For each $\cal S$ the corresponding pair $(\omega,\omega_\varepsilon)$ always lies in the grey shaded area. $\bar\omega_\varepsilon(\omega)$ goes to zero for $\omega\rightarrow 0$ and to $\Omega_\varepsilon$ for $\omega\rightarrow \Omega$. $\bar\omega_\varepsilon(\omega)$ never lies strictly above the straight line from the origin to $(\Omega,\Omega_\varepsilon)$, hence, it is never concave. If it is convex, than Corollary 2 is valid.}
\label{fig:illustration}
\end{figure}

The central question is: Under what conditions does Corollary 2 hold, meaning when do $\phi_\varepsilon({\cal S})$ and $\omega({\cal S})$ not correlate negatively? The proportion of "bad" global minima, $\phi_\varepsilon({\cal S})$, should not have a tendancy of becoming smaller for larger global minima set sizes $\omega({\cal S})$.

We take a look at the random variables $0<\omega({\cal S})\leq \Omega$ and $0\leq \omega_\varepsilon({\cal S})\leq \Omega_\varepsilon$ generated by random drawings of ${\cal S}$ and their joint probability distribution $P(\omega, \omega_\varepsilon)$. With $P(\omega, \omega_\varepsilon)=P(\omega_\varepsilon\vert \omega)P(\omega)$ we can define $\bar \phi_\varepsilon(\omega)$ and $\bar \omega_\varepsilon(\omega)$ according to
\begin{eqnarray*}
    \bar \phi_\varepsilon(\omega)&=&\int_0^{{\Omega}_\varepsilon} \frac{\omega_\varepsilon}{\omega} \, P(\omega_\varepsilon\vert \omega) \, d\omega_\varepsilon \\
    &=&\frac{1}{\omega}\int_0^{{\Omega}_\varepsilon} \omega_\varepsilon \, P(\omega_\varepsilon\vert \omega) \, d\omega_\varepsilon \\
    &=&\frac{\bar \omega_\varepsilon(\omega)}{\omega}.
\end{eqnarray*}

If $\bar \phi_\varepsilon(\omega)$ is monotonically non-decreasing with $\omega$, then
\begin{eqnarray*}
    \left\langle \phi_\varepsilon({\cal S})\omega({\cal S})\right\rangle_{\cal S}-\left\langle \phi_\varepsilon({\cal S})\right\rangle_{\cal S}\left\langle \omega({\cal S})\right\rangle_{\cal S}
    &=&
    \left\langle \phi_\varepsilon \omega\right\rangle_{\omega_\varepsilon,\omega}-\left\langle \phi_\varepsilon\right\rangle_{\omega_\varepsilon,\omega}\left\langle \omega\right\rangle_{\omega_\varepsilon,\omega}\\
    &=&\left\langle \bar \phi_\varepsilon(\omega) \omega\right\rangle_{\omega}-\left\langle \bar \phi_\varepsilon(\omega)\right\rangle_{\omega}\left\langle \omega\right\rangle_{\omega}\\
    &\geq& 0,
\end{eqnarray*}
which is sufficient for Corollary 2 to be valid.

When is this the case? In Fig.\ \ref{fig:illustration} we illustrate some scenarios. With a random ${\cal S}$, the corresponding pair $(\omega, \omega_\varepsilon)$ always falls into the grey area, since always $\omega_\varepsilon\leq \Omega_\varepsilon$ and $\omega_\varepsilon\leq \omega$. For $\omega\rightarrow 0$, $\bar \omega_\varepsilon(\omega)\rightarrow 0$, and for $\omega\rightarrow \Omega$, $\bar \omega_\varepsilon(\omega) \rightarrow \Omega_\varepsilon$. A simple course of $\bar \omega_\varepsilon(\omega)$ between these two points would be convex or concave, in the sense that the first derivative is monotone. However, the case that $\omega_\varepsilon(\omega)$ is strictly above the line given by $\omega_\varepsilon=(\Omega_\varepsilon/\Omega)\, \omega$, which includes being strictly concave, can be ruled out, since otherwise 
\begin{eqnarray*}
    \frac{\langle \omega_\varepsilon({\cal S})\rangle_{\cal S} }{\langle \omega({\cal S})\rangle_{\cal S}}
    &=&\frac{\langle \omega_\varepsilon\rangle_{\omega_\varepsilon,\omega} }{\langle \omega\rangle_{\omega_\varepsilon,\omega}}\\
    &=&\frac{1}{\langle \omega\rangle_{\omega}}\int_{0}^{\Omega} \bar \omega_\varepsilon(\omega)\,P(\omega)d\omega \\
    &>& \frac{1}{\langle \omega\rangle_{\omega}}\int_{0}^{\Omega} \frac{\Omega_\varepsilon}{\Omega}\,\omega\, P(\omega)d\omega \\
    &>& \frac{\Omega_\varepsilon}{\Omega},
\end{eqnarray*}
which contradicts \eqref{eq:leq_one} of Corollary 1. But if $\bar \omega_\varepsilon(\omega)$ is convex, then per definition of convexity
\begin{equation*}
    \frac{\bar \omega_\varepsilon(\omega)}{\omega}=\bar \phi_\varepsilon(\omega)
\end{equation*}
is monotonically non-decreasing in $\omega$ and, hence, Corollary 2 is valid. 

Thus, the trajectory of $\bar \omega_\varepsilon(\omega)$ must at least exhibit a complexity beyond that of a simple convex shape for Corollary 2 to be rendered invalid. While a higher complexity is a plausible scenario, it is not yet sufficient and raises intriguing research questions. Our experiments in the subsequent section consistently affirm the validity of Corollary 2.

\section{Experiments}

To experimentally verify our theoretical results, we study experimental setups in which we can determine $D(E)$, $\omega({\cal S})$ and $\phi_\varepsilon({\cal S})$ by random sampling. We take multi-layer-perceptrons with leaky ReLUs (10\% leakiness) and no bias. The output of the classifier is then given by 
\begin{equation*}
    {\bf y}=ReLU(W_K ReLU(W_{K-1} (...ReLU(W_1{\bf x}))))
\end{equation*}
with $W_k$, $k=1,...,K$ as the weight matrix of layer $k$, ${\bf x}\in \mathbb{R}^D$ as the input vector of dimension $D$ and ${\bf y}\in \mathbb{R}^2$ as the output vector, in our case for binary classification. The ReLU of the output layer with the largest output determines the predicted class membership for the given input ${\bf x}$. In this setting, it is easy to see and well known that the class membership is invariant to the lengths of the weight vectors due to the positive homogeneity of the leaky ReLUs (for $s \geq 0$, $ReLU(s x)=s ReLU(x)$). Thus, each classifier function is determined by a weight vector (comprising all weights of the network) of unit length. ${\cal W}$ is given by the unit sphere in dimension $\mathbb{R}^N$, with $N$ as the number of weights of the network. In our experiments, we will sample uniformly from this unit sphere. With the leakiness of the ReLUs, we avoid too many trivial solutions with many "dead" ReLUs. 

\subsection{Synthetic data set}

For the first experiment we take a synthetic data set of two isotropic Gaussians in 10 dimensions. The Gaussian of class A is centered at $(+1,0, ... ,0)$ and of class B at $(-1,0,...,0)$. Both Gaussians  have a variance of $0.5$ and, hence, overlap with $2.28\%$. This is the minimal possible classification error $E_{min}$. We apply three different networks. The smallest has one hidden layer with ten hidden units, which leads to $120$ weights. The next has again one hidden layer, but $100$ hidden units, hence, with $1,200$ ten times as many weights. The third network has ten hidden layers with ten units each, hence, $1,020$ weights. 

\begin{figure}[t]
	\centering
	\begin{subfigure}[t]{0.32\textwidth}
		\includegraphics[width=1.1\textwidth]{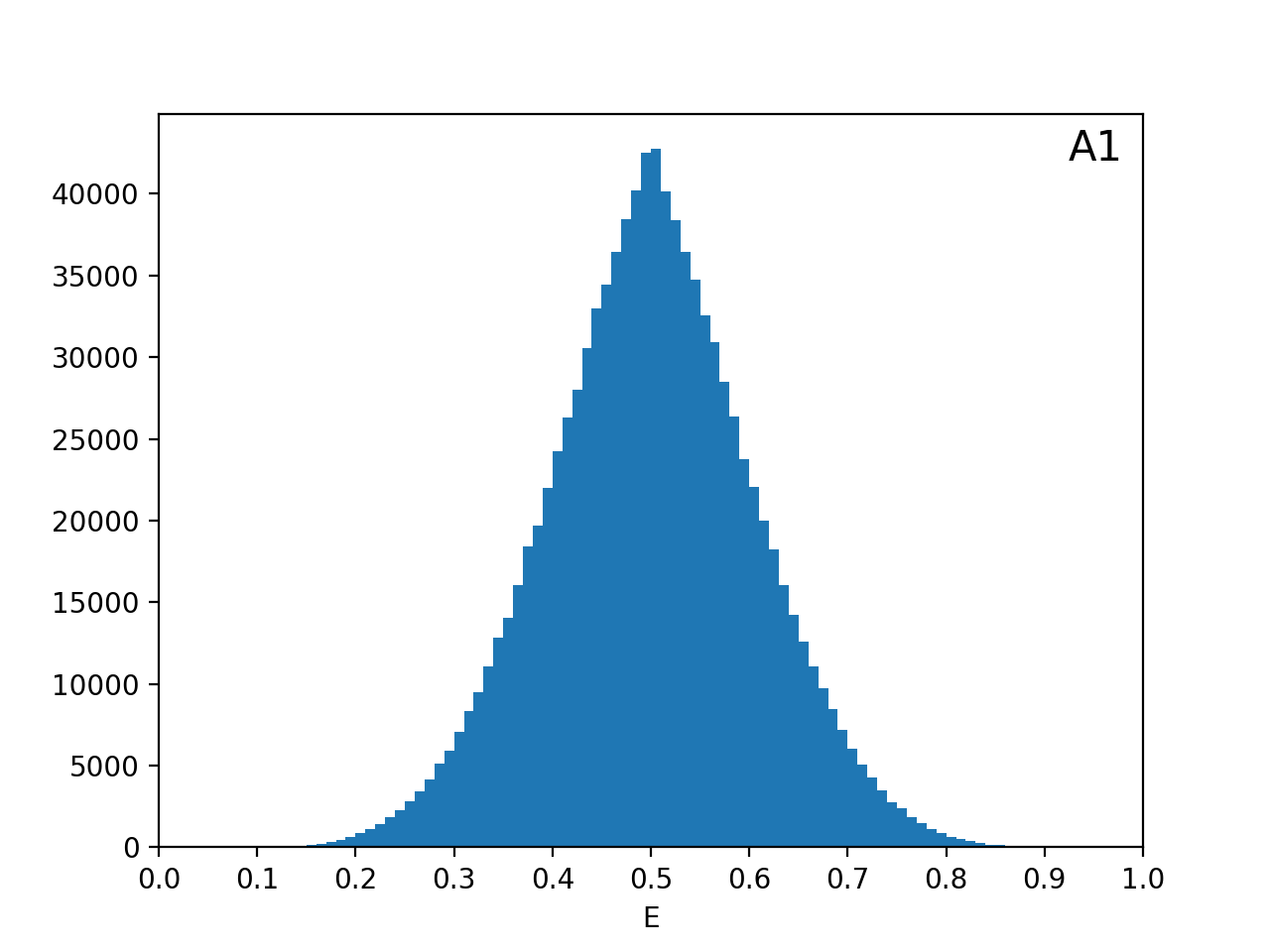}
	\end{subfigure}
    ~
	\begin{subfigure}[t]{0.32\textwidth}
		\includegraphics[width=1.1\textwidth]{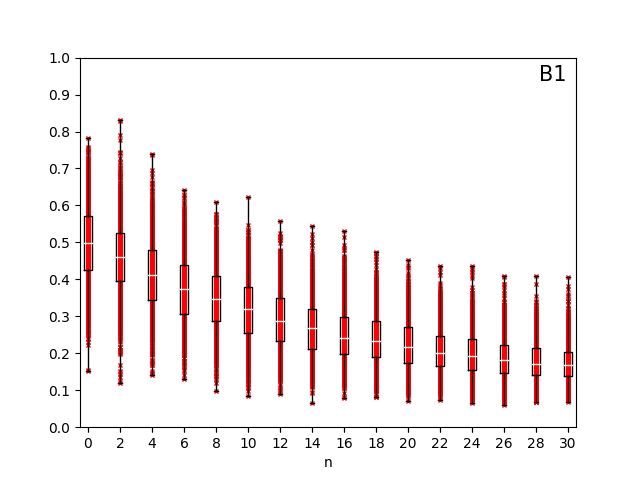}
	\end{subfigure}
    ~
	\begin{subfigure}[t]{0.32\textwidth}
		\includegraphics[width=1.1\textwidth]{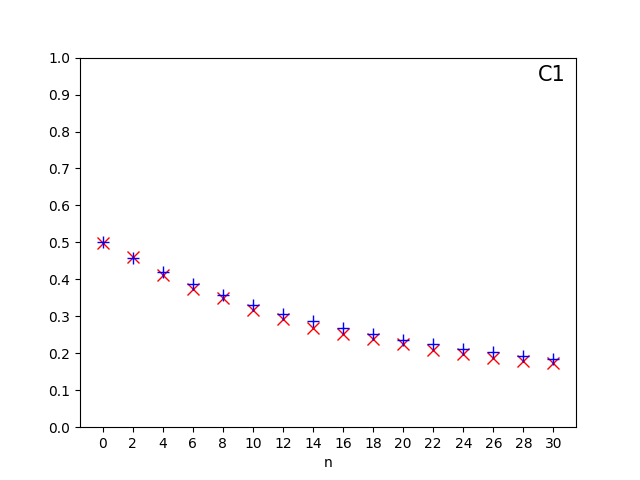}
	\end{subfigure}
	
	\begin{subfigure}[t]{0.32\textwidth}
		\includegraphics[width=1.1\textwidth]{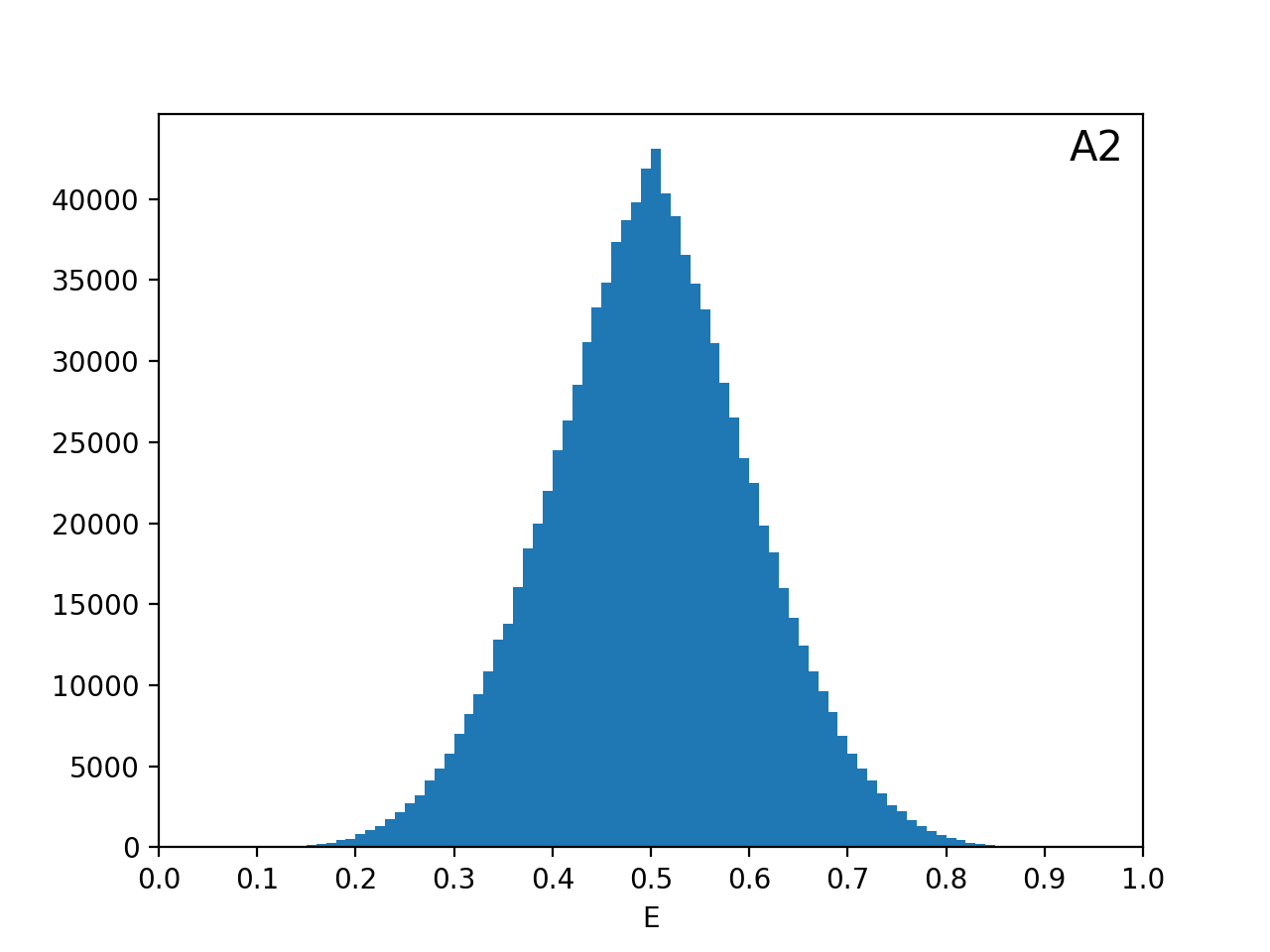}
	\end{subfigure}
    ~
	\begin{subfigure}[t]{0.32\textwidth}
		\includegraphics[width=1.1\textwidth]{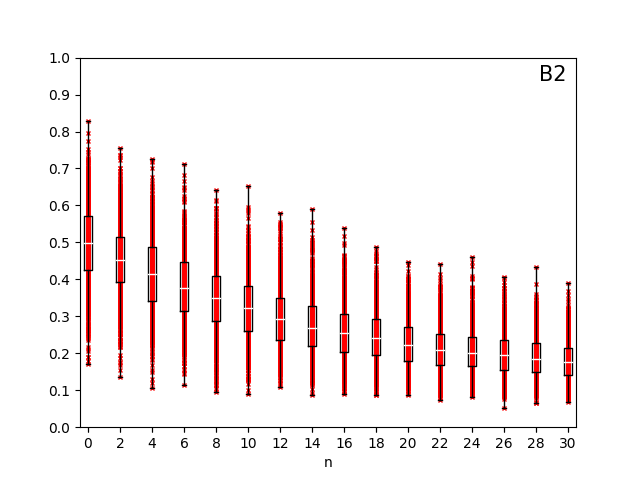}
	\end{subfigure}
   ~
	\begin{subfigure}[t]{0.32\textwidth}
		\includegraphics[width=1.1\textwidth]{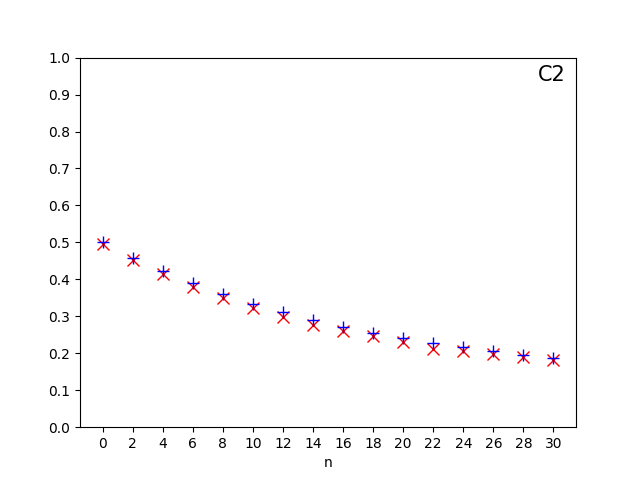}
	\end{subfigure}
	
	\begin{subfigure}[t]{0.32\textwidth}
		\includegraphics[width=1.1\textwidth]{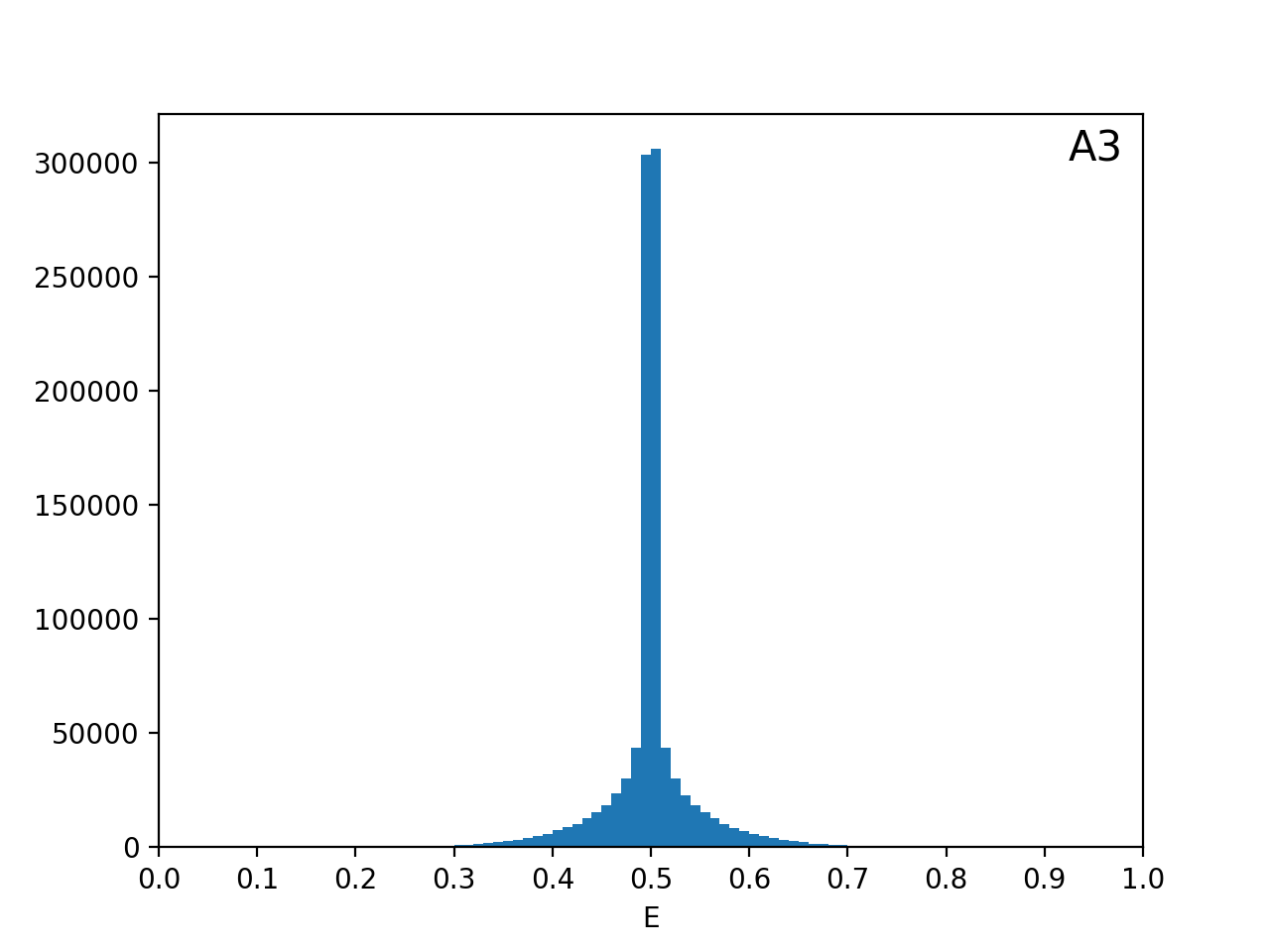}
	\end{subfigure}
    ~
	\begin{subfigure}[t]{0.32\textwidth}
		\includegraphics[width=1.1\textwidth]{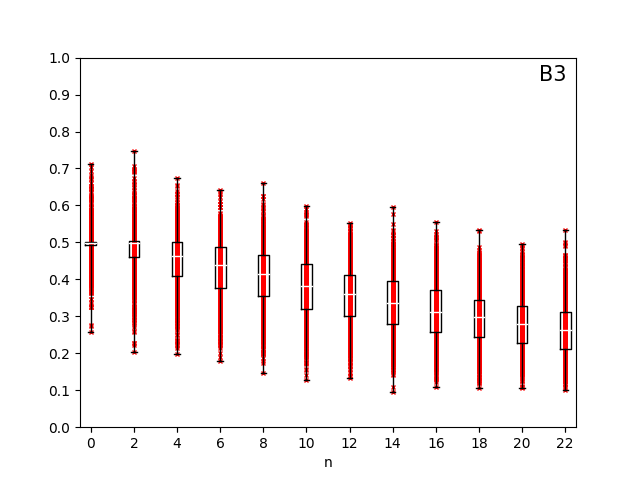}
	\end{subfigure}
    ~
	\begin{subfigure}[t]{0.32\textwidth}
		\includegraphics[width=1.1\textwidth]{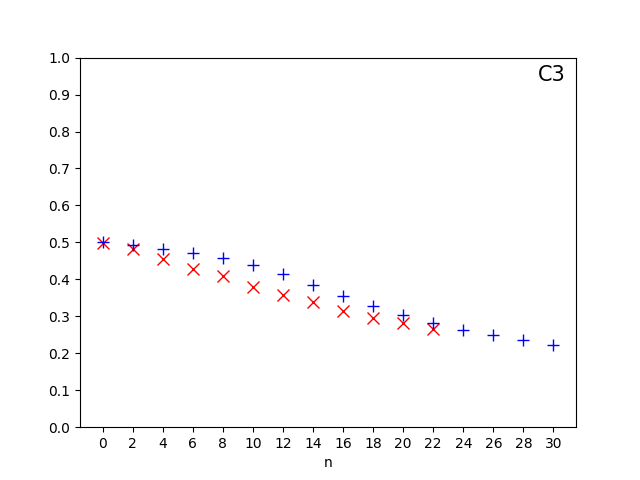}
	\end{subfigure}

	\caption{Experiments with two Gaussians in $10$ dimensions, classified by three different networks from top to bottom. The shape of the density of classifiers $D(E)$ (left column) of the first two networks hardly deviate, in spite of $10$ times more weights in A2 than in A1. The column in the middle shows the distribution of test errors of training error zero solutions for increasing $n$. The average fraction of "bad" solutions converges to zero in these highly over-parameterized scenarios. In the right column the average of the test errors from the middle column (red x) are compared with the values from our bound determined by the $D(E)$s in the left column (blue crosses). In all three cases, the bound is valid, even (almost) exactly in C1 and C2. }
\label{fig:ToyProblem}
\end{figure}

The first row of Fig.~\ref{fig:ToyProblem} shows the results with the smallest network. On the left we see its density of classifiers $D(E)$ as a histogram of $100$ bins (A1). We determined $D(E)$ by randomly sampling weight vectors from a uniform probability distribution on the $120$-dimensional unit hypersphere. For each weight vector the true error of the corresponding classifier is measured with a balanced test set consisting of $10,000$ randomly chosen test data from the two classes. The result is put into the corresponding bin. This was repeated $1,000,000$ times. As expected, $D(E)$ is symmetric and has its peak at $E=0.5$, and good classifiers with a small true error are rare within ${\cal W}$, i.e., on the hypersphere. 

In the middle (B1) we see the distribution of the test (true) errors at randomly found global minima of randomly chosen training sets. For each $n$, the distribution of the test error along the y-axis corresponds to the mean normalized density of global minima $Q_n(E)$. We obtained this by taking a randomly chosen training data set, and then randomly sampled weight vectors from the unit hypersphere until we found one with training error zero on this training set. Then we measured the corresponding true error (red dot) with the test data set and then took the next randomly chosen training data set. For each $n$ this was repeated $1,000$ times ($1,000$ red dots for each $n$). The box-plots show, that already for $n=30$ more than 75\% of the training error zero solutions provide test (true) errors of less than $0.2$, i.e., for $n=30$ we have $\langle \phi_{\varepsilon=0.2}({\cal S})\rangle=0.25$. In only one of the $1,000$ trials there is a test error slightly above $0.4$. "Bad" global minima are quickly becoming rare on the unit hypersphere. 

On the right (C1) we show the average test error (red x) for each $n$, the average over the $1,000$ red dots from (B1). This corresponds to $E_n$ of the l.h.s.\ of \eqref{eq:mean_E}. The r.h.s.\ we can determine with the measured $D(E)$ from the left column, with the integral as sums over the bins. The results are shown by the blue crosses. Red and blue crosses match surprisingly well. Indeed, \eqref{eq:frac_ineq} is valid, there is no negative correlation. Even more, rather equality seems to be given, which means there is no correlation. 

The second row of Fig.~\ref{fig:ToyProblem} shows the results of the network with one hidden layer and $100$ hidden units. This network has ten times as many weights as the previous network with $10$ hidden units. Nevertheless, the shape of the density of classifiers $D(E)$ (A2) hardly deviates from the $D(E)$ of the much smaller network (A1). Also the distribution of the test (true) errors of training errors zero solutions (B1 and B2) are more or less identical. As for the ten times smaller network, already for $n=30$ in about 75\% of the cases one ends up with a test error of less than $0.2$, and in none of the $1,000$ trials there is a test error above $0.4$. Accordingly, also Figs.\ C1 and C2 look identical. Again, the red and blue crosses match in C2 and, hence, equality in \eqref{eq:frac_ineq} is valid and there is no correlation. 

In the third row of Fig.~\ref{fig:ToyProblem} we see the density of classifiers $D(E)$ for the network with ten hidden layers with ten neurons each and $1,020$ weights (A3). The large peak at $E=0.5$ stems from weight configurations where the classifier output is always the same class, independent from the input. The classifiers $h$ which correspond to these weight configurations are ruled out immediately as soon as training data from both classes occur, since then these $h$ do not provide zero training error and do not belong to the global minimum. With $D(E)$ having a different shape as in the two cases before, also Fig.\ B3 looks differently now. It converges more slowly, and the peak of $D(E)$ resolves with increasing $n$, as can be seen from the increasing box. For example, for $n=2$ in half of the training data instances the training data are from the same class and there still are solutions from the peak at $E=0.5$. We stopped with $n=22$, since the time to find zero training error solutions by random sampling started to take too much time. In Fig.\ C3 the red and blue crosses do not match for every $n$, but the blue crosses always lie above the red ones. The bound \eqref{eq:frac_ineq} is valid, but now for some $n$ there is a positive correlation between $\phi_\varepsilon({\cal S})$ and $\omega({\cal S})$.

\begin{figure}[t]
	\centering
	\begin{subfigure}[t]{0.32\textwidth}
		\includegraphics[width=1.1\textwidth]{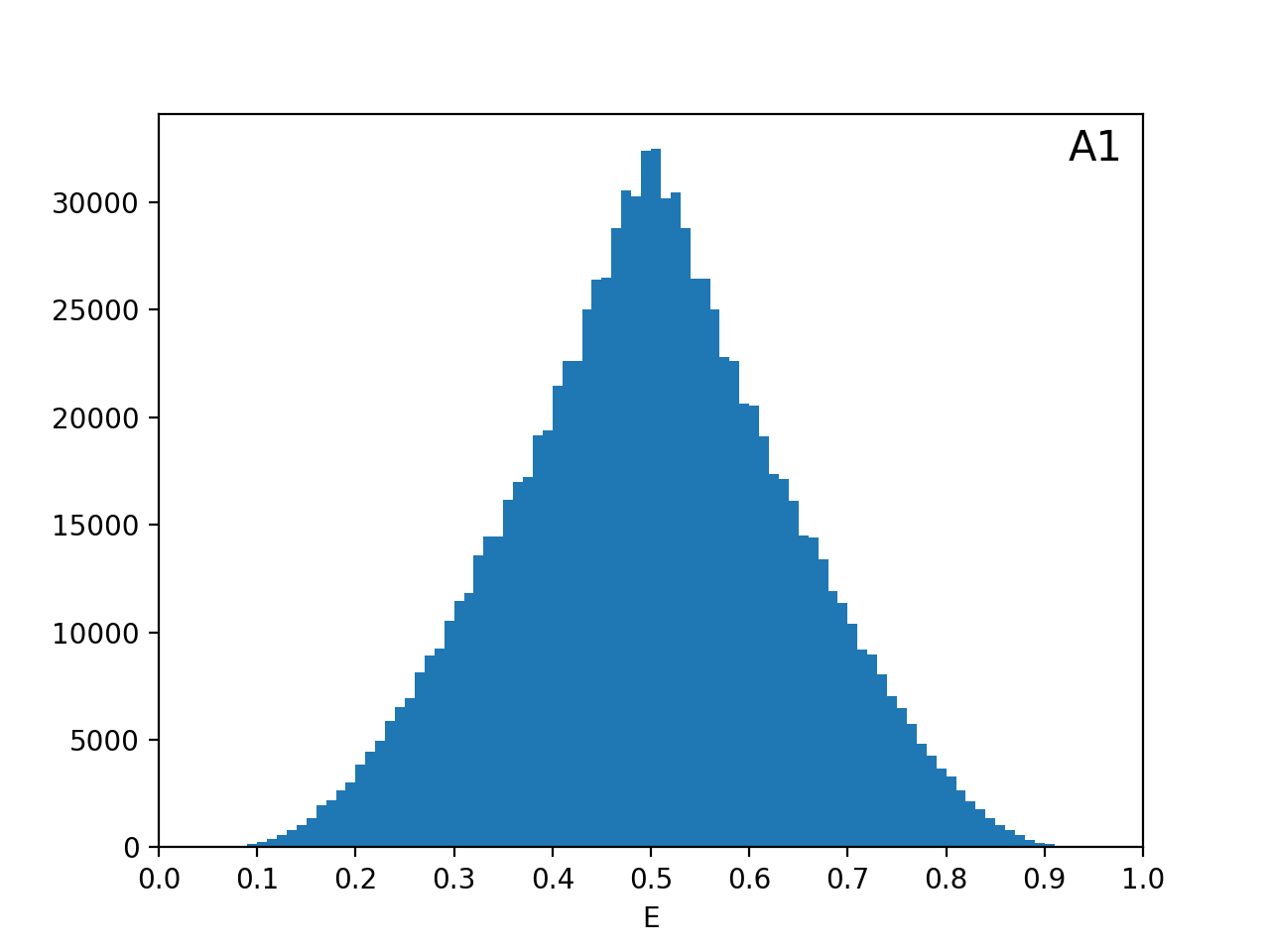}
	\end{subfigure}
    ~
	\begin{subfigure}[t]{0.32\textwidth}
		\includegraphics[width=1.1\textwidth]{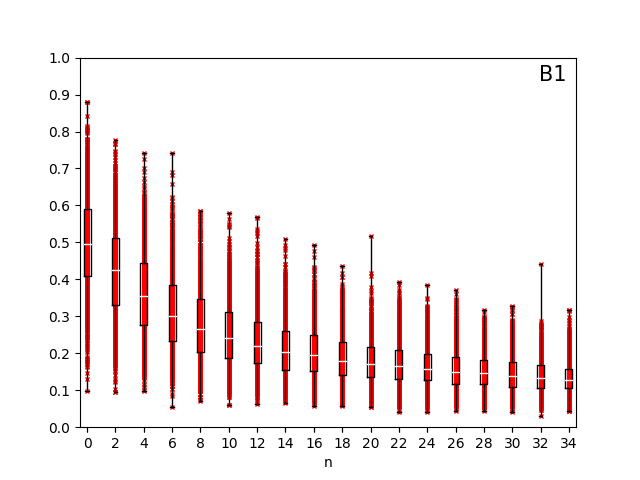}
	\end{subfigure}
    ~
	\begin{subfigure}[t]{0.32\textwidth}
		\includegraphics[width=1.1\textwidth]{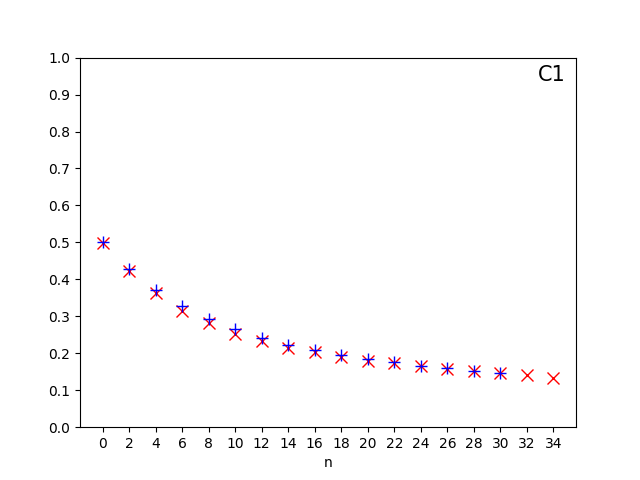}
	\end{subfigure}

	\begin{subfigure}[t]{0.32\textwidth}
		\includegraphics[width=1.1\textwidth]{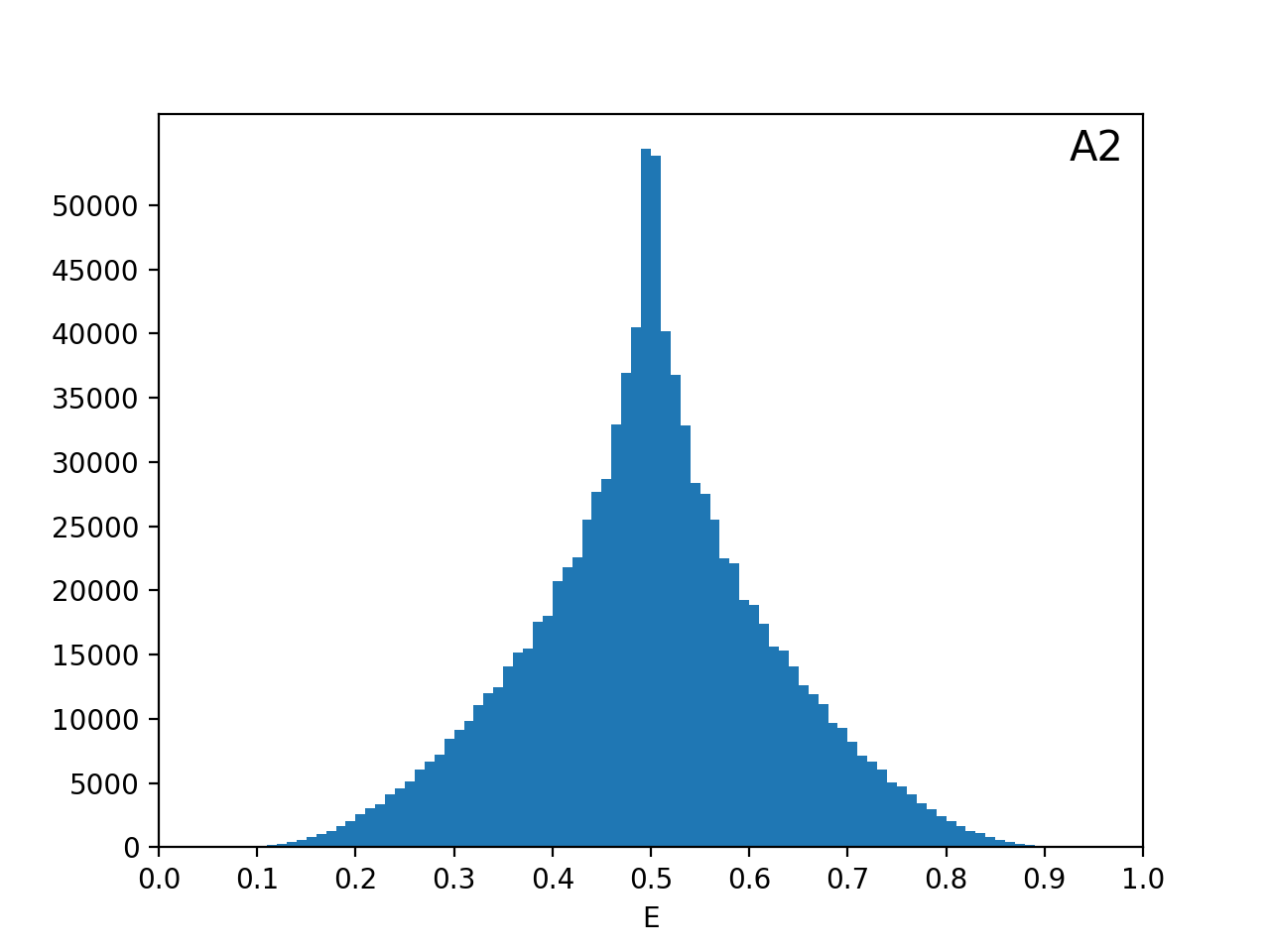}
	\end{subfigure}
    ~
	\begin{subfigure}[t]{0.32\textwidth}
		\includegraphics[width=1.1\textwidth]{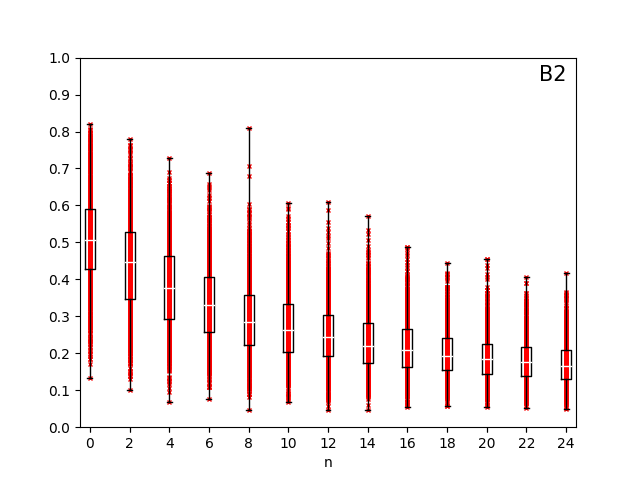}
	\end{subfigure}
    ~
	\begin{subfigure}[t]{0.32\textwidth}
		\includegraphics[width=1.1\textwidth]{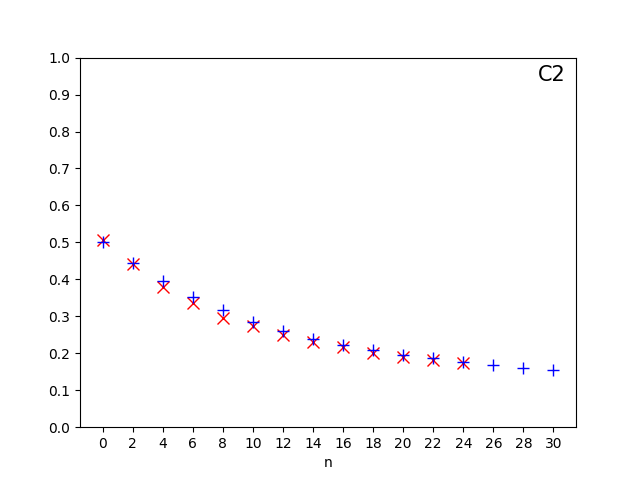}
	\end{subfigure}

	\caption{Classification of the two digits 1 and 2 of the MNIST data set with two output ReLUs (first row) and with one hidden layer with 10 hidden ReLUs and two output ReLUs (second row). Left (A1, A2) the density of classifiers $D(E)$ and in the  middle (B1, B2) the distribution of the test errors (red dots) of training error zero solutions for increasing number of training data $n$. On the right (C1, C2) the average of the test errors (red x) compared with values of the bound (blue crosses) derived from the respective $D(E)$. The bound holds and is (almost) even tight in both cases. }
\label{fig:MNIST}
\end{figure}

\subsection{Data set from MNIST}

In Fig.~\ref{fig:MNIST} we show the results of the same experiments with real data. We took the digits 1 and 2 of the MNIST data set as the two classes to be classified. Each digit image has a size of $28\times 28$ pixels and thus the input dimension is $784$. We used the same network structure as before, in one case with no hidden layer, i.e., only two leaky output ReLUs, one for each class, and in the other case a one-hidden layer MLP with 10 hidden leaky ReLUs and two leaky output ReLUs. In the first case, the classifier has $1,568$ weights, and in the second case $7,860$ weights. The training data sets were randomly drawn from a set of $6,000$ images of each digit, respectively. These images were taken from the MNIST training set. For testing, $900$ images of each digit were taken from the MNIST test set.

The results look very much like in Fig.~\ref{fig:ToyProblem} for the synthetic data. In the first row of Fig.~\ref{fig:MNIST}, the results without hidden layer and $1.568$ weights and in the second row the results with ten hidden units and $7.860$ weights are shown. From left to right we see the density of classifiers $D(E)$, the distribution of test errors at training error zero for different $n$, and the average test errors compared to values of the bound \eqref{eq:frac_ineq}. Again, the bound holds and even equality is given, i.e., there is no correlation between the fraction of "bad" global minima $\phi_\varepsilon({\cal S})$ and the size of the set of global minima $\omega({\cal S})$. 

In both scenarios, the classifiers are highly over-parameterized for these small training data sets, but nevertheless good solutions quickly dominate. For both networks, already for $n=24$ about $75$\% of all test errors are below $0.2$. Instead of a fifty-fifty chance of incorrect predictions, $75$\% of the solutions with zero training error make fewer than one mistake in five cases.

\subsection{VGG and ResNet and Caltech101}

\begin{figure}[t]
	\centering
	\begin{subfigure}[t]{0.48\textwidth}
		\includegraphics[width=1.0\textwidth]{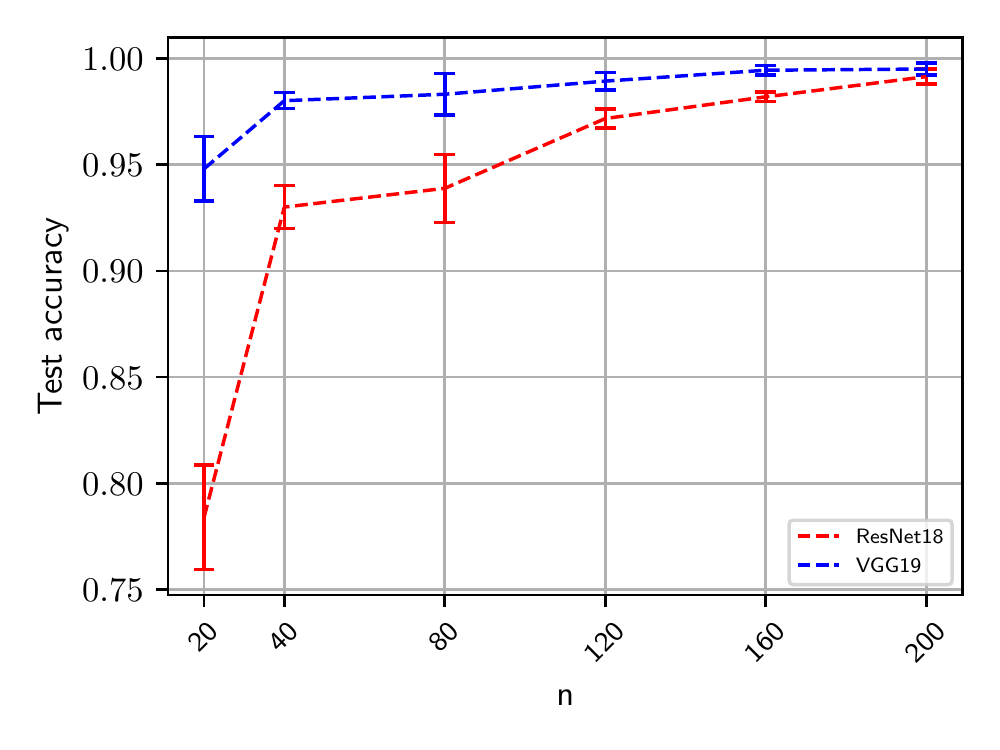}
	\end{subfigure}
    ~
	\begin{subfigure}[t]{0.48\textwidth}
		\includegraphics[width=1.0\textwidth]{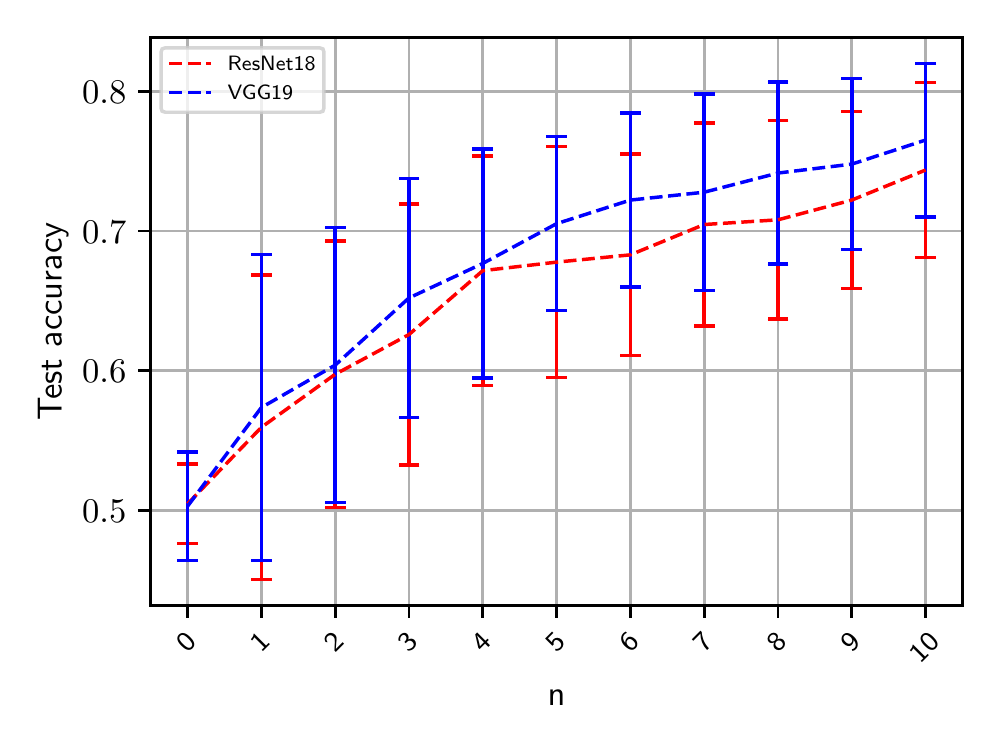}
	\end{subfigure}

	\caption{The left plots are adapted from \cite{linse_large_2023} and show the mean test accuracies (and variances) in classifying motorcycles and airplanes on images from the Caltech101 dataset with VGG19 and ResNet18, respectively. Both networks are trained with stochastic gradient descent up to training error zero. To the right we see the mean test accuracies and their variances of solutions with training error zero obtained by random sampling. Good solutions quickly dominate within the set of solutions with training error zero, and the more than ten times larger VGG19 converges faster than ResNet18, which corresponds to gradient descent.}
\label{fig:Caltech}
\end{figure}

Finally, we take a look at very large Deep Networks. In \cite{linse_large_2023} was shown that a VGG19 net with $140$ million weights is able to learn to discriminate airplanes and motorcycles on images from the Caltech101 dataset \cite{li_andreeto_ranzato_perona_2022} up to $95$\% accuracy trained with only $20$ examples from each class. This extremely over-parameterized network was initialized randomly and then trained by stochastic gradient descent up to training error zero without any regularization. Fig.~\ref{fig:Caltech}A shows these results adapted from \cite{linse_large_2023}. With $200$ training data the test accuracy is almost $100$\% despite complete over-fitting and memorizing the training data. 

We hypothesize that the above results can be explained  by "bad" solutions being rare in the set of training error zero solutions. Therefore, an ERM algorithm like stochastic gradient descent should very likely end up in a good solution also without regularization. We test our hypotheses like in the previous sections. Again, due to the positive homogeneity of the VGG19 net each classifier function realized by VGG19 corresponds to a point on the unit hypersphere in the weight space. We uniformly draw random weights from this sphere, and if the corresponding classifier function has zero training error we determine its test accuracy. The training examples are drawn from 798 motorbike and aircraft images from Caltech101, always the same number for each class, analog to the experiments in Fig.~\ref{fig:Caltech}A. The rest of the images are used for determining the test accuracy. Like in Fig.~\ref{fig:Caltech}A we perform the same experiment also for ResNet18, which has $11,2$ million parameters - more than ten times less than VGG19.    

Fig.~\ref{fig:Caltech}B shows the results for up to $10$ samples per class (for larger $n$ it became too computationally expensive). For each $n$ we randomly drew random training sets and weights until we reached $100$ zero training error solutions. Fig.~\ref{fig:Caltech}B shows the mean of the corresponding test accuracies and their variances. To find zero training error solutions required several million trials for larger $n$. Interestingly, on average it required much more trials for ResNet18 than for VGG19. Already for $n=10$ the training error zero solutions of VGG19 have a mean test accuracy of $75$\%. A rough extrapolation to $n=20$ indeed leads to the range of VGG19 in Fig.~\ref{fig:Caltech}A. Also ResNet18 being worse than VGG19 in spite of being ten times smaller corresponds to Fig.~\ref{fig:Caltech}A. Both is in line with our conjecture.   

\section{Discussion}

Experiments consistently demonstrate that highly over-parameterized classifiers can learn effectively even without explicit regularization and despite "memorizing" the training data up to zero training error. This contradicts traditional machine learning wisdom, rooted in learning theory based on uniform convergence, which typically requires (with high probability) a uniformly small generalization gap. This implies the complete absence of "bad" classifiers within the set of possible ERM solutions with zero training error. This stringent requirement necessitates substantial training data. However, it might be reasonable to allow for a small fraction of "bad" classifiers, rather than demanding zero, within all ERM solutions. As long as the  proportion of "bad" classifiers is small, there remains a high probability for the ERM algorithm to discover a good solution, potentially achieved with significantly less training data.

We proved mathematically that within the over-parameterized regime, the mean set size or parameter volume of "bad" ERM solutions diminishes at a much faster rate with the increase in the number of training samples compared to the mean set size of all possible ERM solutions. If the fraction of "bad" classifiers within the set of ERM solutions does not exhibit a negative correlation with the volume of ERM solutions, the fraction of "bad" classifiers will be small with high probability. 

We introduced the density of classifiers (DOC) at a true error $E$, which depends on the set of classifier functions and the classification problem. The DOC determines the bound for the fraction of "bad" classifiers and serves as the primary characteristic of a given classifier/problem case. Since its shape (i.e., the normalized DOC) is important, it does not necessarily depend on the classifier's capacity or its number of parameters. Increasing the absolute density values by expanding the capacity of the classifier function set may keep the normalized DOC invariant, or may even improve it, as observed in numerous experiments.

We experimentally validated our theoretical results using a synthetic dataset and a subset of MNIST with three and two different network architectures and sizes, respectively. Through random sampling, we determined the density of classifiers (DOC) and the distribution of the true error on the set of ERM solutions. Indeed, the theoretical bounds complied with the experiments, and in most cases the theoretical values even matched precisely with the measured mean errors from the experiments. The experiments have shown that the DOC is not necessarily dependent on the complexity and size of the classifier. Networks with ten times as many parameters had similar DOCs, leading to a comparable decay in the fraction of "bad" minima, consistent with our theoretical findings. Finally, we applied our framework to VGG19 and ResNet18, two Convolutional Neural Networks with millions of parameters. Again, the fraction of good solutions increases immediately already with a handful of training data, in line with the unexpected good generalization of these networks observed in highly over-parameterized settings and without explicit regularization. 

Of course, (stochastic) gradient descent, being the standard Empirical Risk Minimization (ERM) algorithm for learning, does not necessarily converge to each of the global minima with equal probability. It is influenced by various factors such as initialization, error landscape structure, learning rates, and more. Consequently, unlike the random sampling in our experiments, the probability of landing in a "bad" solution is not precisely determined by the fraction of "bad" solutions. Nonetheless, for (stochastic) gradient descent to consistently converge to the tiny fraction of "bad" solutions, it would have to exhibit an "exponential" bias in favor of such solutions. Traditional machine learning wisdom previously assumed the opposite—that good solutions are rare and that substantial guidance, such as explicit or implicit regularization, is essential to locate them and avoid "bad" solutions. In this context, we present a novel perspective that may contribute to understanding the unexpected behavior of highly over-parameterized networks.

In our theoretical framework, we required the absence of a negative correlation between the fraction of "bad" classifiers within the set of global minima and the size of the set of global minima. 
We outlined several considerations concerning the joint distribution of these two quantities and their conditional means, suggesting that the absence of negative correlation appears to be more plausible than the reverse. In our experiments, we indeed observed no negative correlation, and even no correlation at all in most of the evaluated settings. However, since random sampling for finding a training error zero solution becomes very time-consuming with increasing training set size, in our experiments we were restricted to relatively small settings. 

For a comprehensive theoretical understanding, it remains an open question under what conditions, concerning the mathematical structure of the classifier and the classification problem, there exists no negative or even no correlation. Such a correlation is a macroscopic structural characteristic of the classifier/problem setting, which might be influenced by high over-parameterization or specific aspects of the structure of neural networks. This intriguing question remains open for exploration in future research, offering an avenue for deeper insights into the behavior of highly over-parameterized classifiers.

\subsection*{Acknowledgement}

We thank Christoph Linse for discussions and providing the graphs in Figure 5.


\bibliographystyle{ieeetr}
\bibliography{literature}



\newpage
\appendix

\section{Appendix}

\subsection*{Proofs}

{\bf Theorem 1:}
\begin{proof}
For a given $w\in {\cal W}$, the loss $L(y,h_w(x))$ is a binomial random variable assuming the values $0$ or $1$ for training data drawn i.i.d.\ from $P(x,y)$. The probability that $L(y,h_w(x))=0$ for a $(x,y)$ drawn from $P(x,y)$ is $1-E(w)$. Hence, for a ${\cal S}$ randomly drawn from $P(x,y)^n$, the probability for $h_w$ to be a consistent classifier with $L(y_i,h_w(x_i))=0$ for each $(x_i,y_i)\in {\cal S}$ is $(1-E(w))^n$. 

We define the indicator function $\mathbf 1_{\cal S}(h_w)$, which is one for $w\in{\cal W}_\varepsilon({\cal S})$ and otherwise zero. For a fixed $w\in{\cal W_\varepsilon}$ and ${\cal S}$ randomly drawn from $P(x,y)^n$, the probability for $\mathbf 1_{\cal S}(h_w)$ assuming the value one is $(1-E(w))^n$. With this we obtain
\begin{eqnarray}
    \langle \omega_\varepsilon({\cal S})\rangle_{\cal S}&=&\left\langle\int_{{\cal W_\varepsilon}} \mathbf 1_{\cal S}(h_w)\, dw\right\rangle_{{\cal S}} \nonumber\\
    &=&\int_{{\cal W_\varepsilon}} \left\langle\mathbf 1_{\cal S}(h_w)\right\rangle_{{\cal S}}\,dw \nonumber\\
    &=&\int_{{\cal W_\varepsilon}} (1-E(w))^n\,dw \nonumber\\
    &=&\int_{E_{min}+\varepsilon}^1  (1-E)^n \, D(E)dE \nonumber\\
    &\leq& \Omega_\varepsilon (1-(E_{min}+\varepsilon))^n. \nonumber
\end{eqnarray}
With the second step we interchanged integrals. This can be done if for a random sequence ${\cal S}_1,...,{\cal S}_m$ of training sets  
\begin{equation}
    \frac{1}{m}\sum_{i=1}^m \mathbf 1_{{\cal S}_i}(h_w)\quad  \rightarrow \quad \left\langle\mathbf 1_{\cal S}(h_w)\right\rangle_{{\cal S}}\qquad \hbox{uniformly in probability}
\end{equation}
on ${\cal W}_\varepsilon$ for $m \rightarrow \infty$. But this is the case due to the finite VC-dimension of our classifier set ${\cal H}_{\cal W}$. ${\cal H}_{\cal W}$ loses the capacity to shatter ${\cal S}_1, \ldots, {\cal S}_m$ with $m$ becoming sufficiently large and, hence, the Uniform Convergence Theorem \cite{VapnikChernovenkis1971} applies. 

\end{proof}

{\bf Corollary 1:}
\begin{proof}
    We show that for any $a>1$ and $0\leq\varepsilon< 1-E_{min}$ 
    \begin{equation*}
        \frac{\langle \omega_\varepsilon({\cal S})\rangle_{\cal S}}{\langle \omega({\cal S})\rangle_{\cal S}}
            \leq \frac{\Omega_\varepsilon}{\Omega}\frac{1}{(1-g_{\varepsilon/a})+g_{\varepsilon/a}\;e^{\left(1-1/a\right)\varepsilon n}}\leq \frac{1}{g_{\varepsilon/a}}\;e^{-\left(1-1/a\right)\varepsilon n} .
    \end{equation*}
    Corollary 1 is the special case for $a=2$. The r.h.s.\ of course requires $g_{\varepsilon/a}>0$.
    
    With Theorem 1 and dividing  the integral in the denominator into two parts we have
    \begin{eqnarray*}
        \frac{\langle \omega_\varepsilon({\cal S})\rangle_{\cal S} }{\langle \omega({\cal S})\rangle_{\cal S}}&=&\frac{\int_{E_{min}+\varepsilon}^1(1-E)^n D(E)\, dE}{\int_0^1(1-E)^n D(E)\, dE}\\
        &=&\frac{\int_{E_{min}+\varepsilon}^1(1-E)^n D(E)\, dE}{\int_0^{E_{min}+\varepsilon}(1-E)^n D(E)\, dE+\int_{E_{min}+\varepsilon}^1(1-E)^n D(E)\, dE}\\
        &=&\frac{1}{1+\frac{\int_0^{E_{min}+\varepsilon}(1-E)^n D(E)\, dE}{\int_{E_{min}+\varepsilon}^1(1-E)^n D(E)\, dE}}\\
        &\leq&\frac{1}{1+\frac{\int_0^{E_{min}+\varepsilon}(1-E)^n D(E)\, dE}{\Omega_\varepsilon(1-(E_{min}+\varepsilon))^n}}
    \end{eqnarray*}
    For any $a>1$ and dividing the integral into two parts, we obtain for the nominator 
    \begin{eqnarray*}
       \int_0^{E_{min}+\varepsilon}(1-E)^n D(E)\, dE
       &=&\int_0^{E_{min}+\varepsilon/a}(1-E)^n D(E)\, dE+\int_{E_{min}+\varepsilon/a}^{E_{min}+\varepsilon}(1-E)^n D(E)\, dE\\
        &\geq&(\Omega-\Omega_{\varepsilon/a})(1-(E_{min}+\varepsilon/a))^n+(\Omega_{\varepsilon/a}-\Omega_\varepsilon)(1-(E_{min}+\varepsilon))^n.
    \end{eqnarray*}
    But then with some basic transformations
    \begin{eqnarray*}
        \frac{\langle \omega_\varepsilon({\cal S})\rangle}{\langle \omega({\cal S})\rangle}
        &\leq&\frac{1}{1+\frac{\Omega_{\varepsilon/a}-\Omega_\varepsilon}{\Omega_\varepsilon}+\frac{(\Omega-\Omega_{\varepsilon/a})(1-(E_{min}+\varepsilon/a))^n}{\Omega_\varepsilon(1-(E_{min}+\varepsilon))^n}}\\
        &\leq&\frac{\Omega_\varepsilon}{\Omega}\frac{1}{(1-g_{\varepsilon/a})+g_{\varepsilon/a}\;\frac{(1-(E_{min}+\varepsilon/a))^n}{(1-(E_{min}+\varepsilon))^n}}.
    \end{eqnarray*}
    For the factor in the denominator with $n$ in the exponent we can write 
    \begin{eqnarray*}
        \frac{(1-(E_{min}+\varepsilon/a))^n}{(1-(E_{min}+\varepsilon))^n}
        &=&e^{n(\ln(1-(E_{min}+\varepsilon/a))-\ln(1-(E_{min}+\varepsilon)))}\\
        &\geq&e^{\left(1-1/a\right)\varepsilon n}.
    \end{eqnarray*}
        The last step uses the Taylor-expansion $\ln(1-x)=-x-x^2/2-x^3/3-...$. With this expansion it is easy to see that
    \begin{equation*}
        \ln(1-(E_{min}+\varepsilon/a))-\ln(1-(E_{min}+\varepsilon))\geq (1-1/a)\varepsilon,
    \end{equation*}
    which concludes the proof.
\end{proof}

\end{document}